\pdfoutput=1
\documentclass[runningheads]{llncs}
\usepackage[table]{xcolor}
\usepackage{graphicx}

\usepackage{tikz}
\usepackage{comment}
\usepackage{amsmath,amssymb} 
\usepackage{color}

\usepackage[accsupp]{axessibility}  

\usepackage{hyperref}
\hypersetup{pdfauthor={Theodoridis},pdftitle={Trapped in texture bias? A large scale comparison of deep instance segmentation}}
\usepackage{url}
\usepackage{pifont}
\usepackage{float}

\begin{document}
\pagestyle{headings}
\mainmatter
\def\ECCVSubNumber{7815}  

\title{Trapped in texture bias? A large scale comparison of deep instance segmentation}

\titlerunning{A large scale comparison of deep instance segmentation}
%
\author{Johannes Theodoridis\inst{1,2}\and
Jessica Hofmann\inst{1}\and
Johannes Maucher\inst{1}\and
\mbox{Andreas Schilling}\inst{2}}
\authorrunning{J. Theodoridis et al.}
%
\institute{
Institute for Applied AI - Hochschule der Medien Stuttgart, Germany
\email{\{theodoridis,jh275,maucher\}@hdm-stuttgart.de}
\and
University of Tübingen, Germany
\email{andreas.schilling@uni-tuebingen.de}
}
\maketitle

\begin{abstract}
Do deep learning models for instance segmentation \mbox{generalize} to novel objects in a systematic way? For classification, such \mbox{behavior} has been questioned. In this study, we aim to understand if certain design decisions such as \textit{framework}, \textit{architecture} or \textit{pre-training} contribute to the semantic understanding of instance segmentation. To answer this question, we consider a special case of robustness and compare pre-trained models on a challenging benchmark for object-centric, out-of-distribution texture. We do not introduce another method in this work. Instead, we take a step back and evaluate a broad range of existing \mbox{literature}. This includes Cascade and Mask R-CNN, Swin \mbox{Transformer}, BMask, YOLACT(++), DETR, BCNet, SOTR and SOLOv2. We find that YOLACT++, SOTR and SOLOv2 are significantly more robust to out-of-distribution texture than other frameworks. In addition, we show that deeper and dynamic architectures improve robustness whereas training schedules, data augmentation and pre-training have only a minor impact. In summary we evaluate 68 models on 61 versions of \mbox{MS COCO} for a total of 4148 evaluations.\newline

Code: \url{https://github.com/JohannesTheo/trapped-in-texture-bias}

\keywords{robust vision, instance segmentation, deep learning, object-centric, out-of-distribution, texture robustness}
\end{abstract}

\section{Introduction}

In this study, we investigate a special case of robustness for deep learning based instance segmentation. More precisely, we want to learn how pre-trained models compare in the case of \textit{out-of-distribution texture}, i.e. when learned objects contain textures that do not appear in the training data. In particular, we aim to understand if different \textit{frameworks}, \textit{architectures} and \textit{pre-training schemes} contribute to model robustness in a systematic way. Despite their remarkable success in computer vision, deep neural networks still struggle in many challenging real-world scenarios \cite{yuille_deep_2018,michaelis2019dragon,recht_imagenet_2019,madan_small_2021}. One specific example are naturally adversarial objects \cite{lau_natural_2021}. Consider for instance a pedestrian with an unconventionally textured dress or a rare horse statue made out of bronze. The model might have seen many pedestrians or natural horses during training but still fails to detect these rare or unseen examples, often with high confidence. Generalizing to such instances is typically described as \textit{out-of-distribution robustness} \cite{hendrycks_natural_2021,lau_natural_2021}. For classification, \cite{hendrycks_natural_2021} suggest that improvements in this direction are more likely to come from computer vision architectures than from existing data augmentation or additional public datasets. Motivated by these findings, we take a step back and perform an extensive comparison of existing literature. Since instance segmentation methods are quite complex, our goal is to unveil the impact of different components and design decision. As a result, our systematic baseline enables more informed design decisions regarding segmentation robustness in the future.
\newline

\begin{figure}[t]
   \begin{center}
   \includegraphics[trim=0 35 0 0, width=\textwidth]{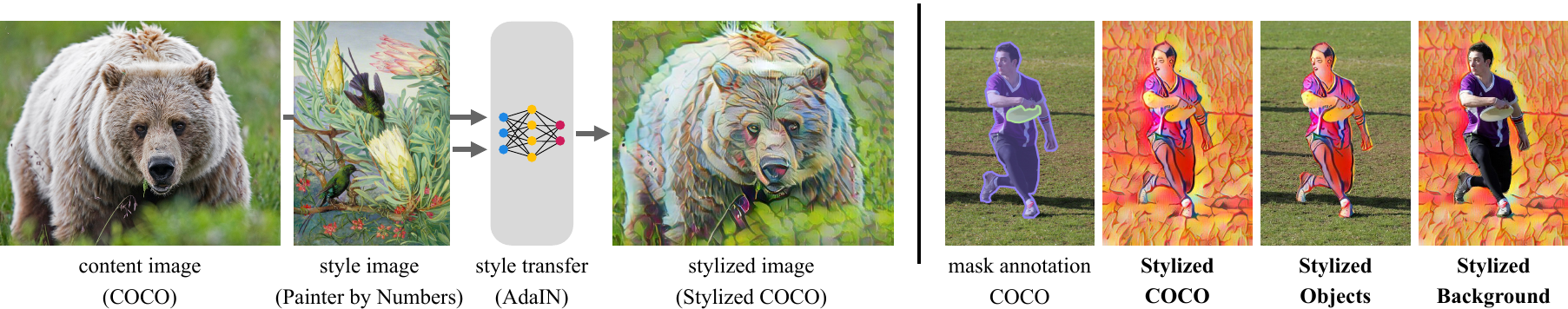}
   \end{center}
   \caption{Left: Simplified creation process of the Stylized COCO dataset. Style images are randomly chosen from Kaggles Painter by Numbers dataset. Right: We use mask annotations to create counterfactual, object-centric versions of Stylized COCO. We include more examples of the creation process in the supplementary material}
   \label{fig:dataset-introduction}
\end{figure}

The problem space we consider is inspired by work from \cite{baker_deep_2018} and \cite{geirhos_imagenet-trained_2019} on texture bias in convolutional neural networks (CNNs). Both groups found that when compared to humans, CNNs for classification ignore object shape in favor of local texture cues. In fact, \cite{brendel_approximating_2019} have further shown that CNNs can robustly classify objects in texturized images where the global appearance of objects is fully mixed up. Since human vision is fairly robust to novel texture and sensitive to object shape, we hypothesize that segmentation models with a similar bias will generalize in more systematic ways as well. As a first step in this direction, we want to learn if existing methods may contain components and design decisions that promote such behavior. In consequence, we opted to evaluate an extensive range of pre-trained models on a challenging but easy to understand edge case.
\newline

As shown in \autoref{fig:dataset-introduction} left, we utilize the AdaIN method \cite{huang_arbitrary_2017} to create a stylized version of MS COCO \cite{fleet_microsoft_2014}. The resulting dataset can be understood as a simulation of familiar objects with guaranteed novel texture, i.e. out-of-distribution texture. Crucially, it ensures that potentially confounding biases from the original data, such as class imbalance or specific view points, are preserved. The simulation is not perfect however. It introduces processing artifacts which we analyze in depth in our methods section. In addition, we control the strength of the style transfer and report results on the full range. This step is essential to distinguish between the effect of image corruption and actually novel object texture. In this sense, style transfer as a whole can be understood as a special type of image corruption. The statistical difference to classical corruption types, such as gaussian noise, is that the latter assumes the corruption to be independent from the signal. In style transfer, the \textit{corruption} is by design not random and highly correlated with the shape and texture features of the content and style image respectively. Alternatives to our simulation approach are discussed in the related work part.
\newline

The complete benchmark setting is displayed in \autoref{fig:dataset-introduction} right. As can be seen, we utilize the segmentation labels to create two additional object-centric versions of Stylized COCO. The motivation for this step is twofold. First, the masking ensures that object contour is recovered in cases where strong stylization results in a camouflage setting. Second and more importantly, it is more plausible to limit the texture simulation to actual object instances instead of the full image. In addition to this causal justification, we are interested to see if pre-trained models are able to exploit this \textit{implicit encoding} of ground truth information which is trivial to spot for the human observer. The Stylized Background dataset can be seen as a control group that allows us to measure the importance of context information. In general, we expect all pre-trained models to degrade with increasing out-of-distribution texture. The approach we take is therefore a negative test, i.e. if some models degrade significantly less than others we consider them more robust.
\newline

\section{Methods}

In this section we describe the datasets, frameworks and models that are used in this study. The code to reproduce our results and the resulting detection and evaluation data can be found here: \url{https://github.com/JohannesTheo/trapped-in-texture-bias}.

\subsection{An object-centric version of Stylized COCO}

\begin{figure}[t]
   \begin{center}
   \includegraphics[trim=0 5 0 0, width=0.8\textwidth]{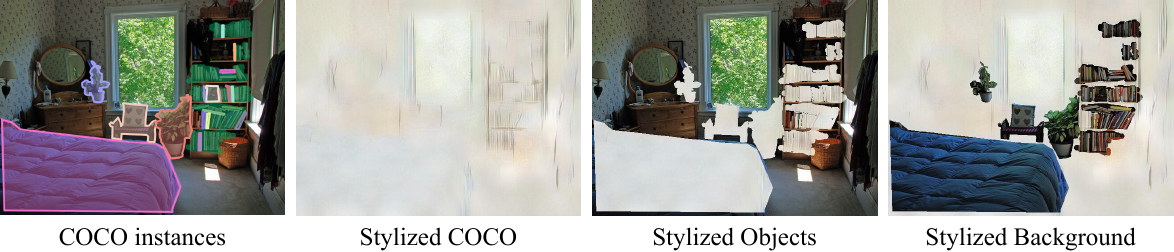}
   \caption{Depending on the style image, object boundaries can vanish due to strong stylization. The Stylized Objects and Background versions of Stylized COCO resolve this issue}
   \label{fig:strong-style}
   \end{center}
\end{figure}

Stylized COCO as shown in \autoref{fig:dataset-introduction} left is an adaptation of Stylized-ImageNet by \cite{geirhos_imagenet-trained_2019}. It was first used by \cite{michaelis2019dragon}\footnote{\tiny Stylized Datasets: \url{https://github.com/bethgelab/stylize-datasets}} as data augmentation during training to improve the robustness of detection models against common corruption types, e.g. gaussian noise or motion blur. We instead use a stylized version of the \texttt{val2017} subset to test pre-trained instance segmentation models directly on this data. By manual inspection of Stylized COCO, we found that strong stylization can sometimes lead to images where the object contour starts to vanish, up to the point where objects and their boundaries dissolve completely. This effect depends on the style image but affects objects of all scales alike. As shown in \autoref{fig:strong-style}, we resolve this issue by using the ground truth mask annotations to limit the style transfer to the actual objects or the background. This not only ensures that object contours are preserved but also controls for global stylization as a confounding variable. By assuming an object-centric causal model, Stylized COCO allows us to ask interventional questions regarding the original COCO dataset, e.g. \textit{``What happens if we change the texture of images?''}. By masking the style transfer to objects or background, we can also ask counterfactual questions such as \textit{``Was it actually the object that caused the change in performance?''}, \textit{``What if we change the background instead?''}. We will refer to the different dataset versions as Stylized COCO (\ding{108}), Stylized Objects (\ding{115}) and Stylized Background (\ding{110}).
\newline

\begin{figure}[t]
\begin{center}
\includegraphics[trim=0 25 0 0, width=\textwidth]{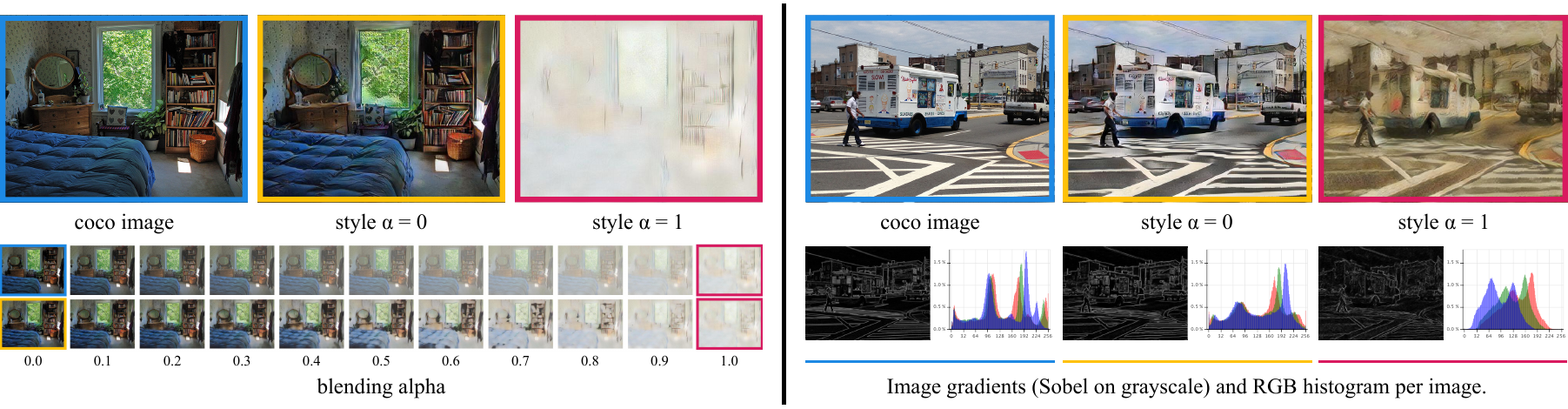}
\end{center}
\caption{Top row: Comparison of COCO and Stylized COCO at different alphas. The AdaIN method introduces subtle artifacts even at \(\alpha=0\) (no style). Bottom left: We control the style strength in feature space (yellow to pink) and pixel space (blue to pink). Every alpha depicts a complete version of the accordingly styled \texttt{val2017} subset. Bottom right: Comparison of image gradients and color histograms at different alphas}
\label{fig:dataset-details}
\end{figure}

A second problem that remains is that shape information within the object can also be lost due to strong stylization. We address this issue by controlling the strength of the AdaIN method. This can be done with an \(\alpha\) parameter that acts as a mixing coefficient between the content and style image. More precisely, AdaIN employs a pre-trained VGG encoder \(f\) on both images, performs an interpolation step between the resulting feature maps and produces the final output with a learned decoder network \(g\). In summary, a stylized image \(t\) is produced by
\begin{equation}
T(c,s,\alpha) = g((1 - \alpha)f(c) + \alpha \text{AdaIN}(f(c),f(s)))
\end{equation}
where \(c\) and \(s\) are the content and style images respectively. We will refer to this method as blending in \textit{feature space}. The top row of \autoref{fig:dataset-details} shows two examples of the extreme points \(\alpha=0\) (no style) and \(\alpha=1\) (full style). Note that at \(\alpha=0\), the image colors are mostly preserved but the algorithm has already introduced artifacts in the form of subtle texture and shape changes. In response, we create a control group where we perform alpha blending between the pixel values of the original content image \(c\) and the stylized image \(t\) at a specific alpha value:
\begin{equation}
P(c,t_{\alpha},\alpha) = (1 - \alpha) * c + \alpha * t_{\alpha} 
\end{equation}
We will refer to this method as blending in \textit{pixel space}. In contrast to the feature space sequence, the control group should preserve textures and object shape over a longer range. The idea is to compare models on both sequences in order to attribute performance to either image corruption or actual out-of-distribution texture. In contrast to \cite{geirhos_imagenet-trained_2019} who used a fixed style strength to modify ImageNet features (\(\alpha=1\)), we produce the full alpha-range \(\alpha \in (0.0, 0.1, 0.2, ..., 1.0)\) for both blending spaces. Note that every alpha value depicts a separate and complete copy of the accordingly styled COCO \texttt{val2017} subset. The qualitative differences can be inspected in \autoref{fig:dataset-details} bottom left (zoom in for better \mbox{visibility}).

\begin{figure}[]
   \begin{center}
      \includegraphics[trim=0 0 40 0, width=0.75\textwidth]{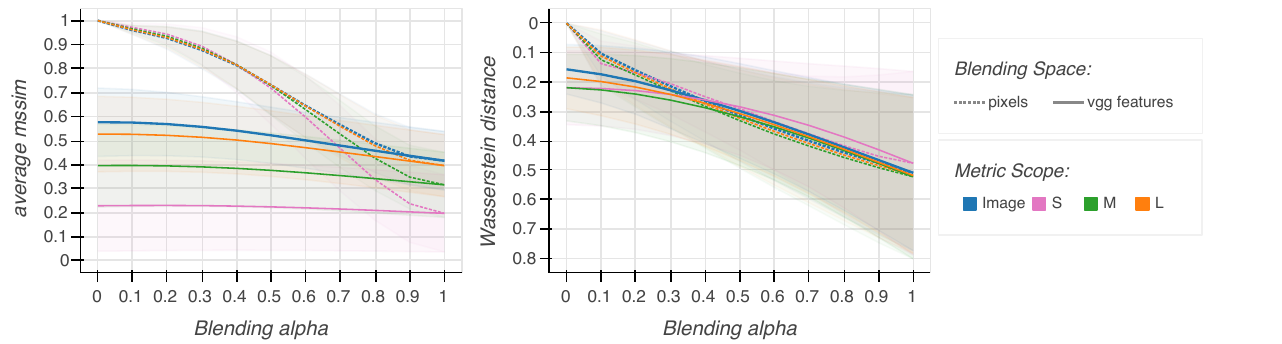}
      \caption{Left: Average structural similarity between image gradients in relation to COCO (a score of 1 means that there is no difference between images). Right: Wasserstein distance between RGB histograms (reversed y-axis)}
      \label{fig:distance-metrics}
   \end{center}
\end{figure}

\subsubsection{Quantitative measures} have been calculated to validate our subjective impression of Stylized COCO. \mbox{\autoref{fig:dataset-details}} bottom right shows a comparison of image gradients and RGB histograms at the \textit{extreme} points. Compared to the original image we can observe the subtle shape changes in the gradient map of \(\alpha=0\) and a significantly different color histogram at \(\alpha=1\). To describe this effect over the full alpha range, we compute the structural similarity index (SSIM) \cite{wang_image_2004} between gradient images of corresponding image pairs. Between RGB histograms we compute the Wasserstein distance alike. We always compare against the original COCO data and report the mean distance averaged over the full dataset at a specific alpha. In addition to the image-to-image scores we also include an instance level comparison for the COCO scales S,M and L. Instances have been cropped based on bbox information. This step was added after we observed that small objects appear to be more affected by the AdaIN artifacts compared to medium and large instances. \autoref{fig:distance-metrics} displays the results and confirms our assumption. Structural similarity depends on object size and is in fact, almost constant over the full feature space range for small objects. Furthermore, the control group preserves structural similarity over a longer range as intended. Color distance in contrast converges at around \(\alpha=0.3\). Based on these insights, we feel confident to better attribute potential performance dips to either image corruption or out-of-distribution texture and subsequently, determine the relative importance of each feature type.

\subsection{Model Selection}

To contribute a comprehensive overview on model robustness, we opted for a broad comparison of popular frameworks and architectures. The dimensions we consider to be impactful are \textit{framework}, \textit{architecture} and \textit{pre-training}. The finally selected models can be found in \autoref{tab:model-overview}.

\subsubsection{Frameworks} for instance segmentation can be categorized in different ways. A first distinction can be made between methods that solve the detection problem as a refinement process of box proposals (multi stage) and methods that predict bounding boxes directly (one stage). We include the popular multi-stage frameworks Mask R-CNN \cite{he_mask_2017} and Cascade Mask R-CNN \cite{cai_cascade_2018} that uses multiple refinement stages instead of one. Both frameworks formulate instance segmentation as a pixel-wise classification problem. Since this rather naive extension to Faster R-CNN \cite{ren_faster_2015-1} can ignore object boundaries and shapes, we include the boundary-preserving mask head alternative (Cascade-) BMask \cite{vedaldi_boundary-preserving_2020} for comparison. A remaining challenge to boundary detection are overlapping objects that occlude the ground truth contour of other instances. We therefore include the Bilayer Convolutional Network (BCNet) \cite{ke_deep_2021} as another mask head alternative. In BCNet, the occluded and occluding objects are detected separately and modeled explicitly in a layered representation. The mask head can then "consider the interaction between [the decoupled boundaries] during mask regression \cite{ke_deep_2021}." A second distinction between frameworks concerns the use of predefined anchor boxes. Anchor based methods predict relative transformations on these priors whereas anchor free methods predict absolute bounding boxes. We include YOLACT(++) \cite{bolya_yolact_2019,bolya_yolact_2020} as a one-stage, anchor based framework. YOLACT is a real-time method that solves instance segmentation without explicit localization (feature pooling). Instead, it generates prototype masks over the entire image which are combined with per-instance mask coefficients to form the final output. The (++) version improves by adding a mask re-scoring branch \cite{huang_mask_2019} and deformable convolutions (v2) \cite{dai_deformable_2017,zhu_deformable_2019}. We include DETR \cite{vedaldi_end--end_2020} as a one-stage, anchor free framework that formulates object detection as a set prediction problem over image features. Note that it was not primarily designed for instance segmentation but offers a corresponding extension that we use in our study. Based on model availability we include BCNet in the FCOS \cite{Tian_2019_ICCV} variant (F-BCNet). FCOS is a fully convolutional, one-stage, anchor-free alternative to Faster R-CNN that "solves object detection in a per-pixel prediction fashion, analogue to semantic segmentation \cite{Tian_2019_ICCV}." Finally, we distinguish  between top down frameworks where detection precedes segmentation and bottom up methods where bounding boxes are derived from mask predictions. We include the bottom-up methods SOLOv2 \cite{wang_solov2_2020} and SOTR \cite{guo_sotr_2021}. SOLO \cite{vedaldi_solo_2020} divides the input into a fixed grid and predicts a semantic category and corresponding instance mask at each location. The final segmentation is obtained with non-maximum-suppression on the gathered grid results to resolve similar predictions of adjacent grid cells. SOLOv2 improves by introducing dynamic convolutions to the mask prediction branch, i.e. an additional input dependent branch that dynamically predicts the convolution kernel weights. A similar idea was used by \cite{vedaldi_conditional_2020}. SOTR uses a twin attention mechanism \cite{huang_ccnet_2019} to model global and semantic dependencies between encoded image patches. The final result is obtained by patch wise classification and a multi-level upsampling module with dynamic convolution kernels for mask predictions, similar to SOLOv2. For completeness, we also include YOLO(v3,4 and scaled v4) to our comparison since detection is a vital sub-task of top down frameworks \cite{redmon_yolov3_2018,bochkovskiy_yolov4_2020,wang_scaled-yolov4_2021}.

\subsubsection{Architectures} used in instance segmentation can be divided into backbone, neck and functional heads. The latter output the final results and are framework specific. Backbones and necks however are typically chosen from a pool of established models which allows for a controlled comparison. The role of backbone networks is to extract meaningful feature representations from the input, i.e. to encode the input. The neck modules define which representations are available to the functional heads, i.e. define the information flow. We include the CNN backbones ResNet \cite{he_deep_2016}, ResNext \cite{xie_aggregated_2017} and RegNet \cite{radosavovic_designing_2020}, a network found with meta architecture search that outperforms EfficientNet \cite{tan_efficientnet_2019}. Note that BCNet utilizes a Graph Convolutional Network (GCN) \cite{kipf_semi-supervised_2017} within its mask heads to model long-range dependencies between pixels (to evade local occlusion). Furthermore, DETR and SOTR are hybrid frameworks that use transformer architectures to process the encoded backbone features. With Swin Transformer \cite{liu_swin_2021} we also include a convolution free backbone alternative based on the Vision Transformer approach (ViT) \cite{dosovitskiy_image_2021}. The most popular neck choice is the Feature Pyramid Network (FPN) \cite{lin_feature_2017}. It builds a hierarchical feature representation from intermediate layers to improve performance at different scales, e.g. small objects. For comparison we also include a ResNet conv4 neck (C4) as used in \cite{ren_faster_2015-1} and a ResNet conv5 neck with dilated convolution (DC5) as used by \cite{dai_deformable_2017}. Finally we abbreviate FPN models that use deformable convolutions as DCN \cite{dai_deformable_2017,zhu_deformable_2019}. Similar to dynamic convolutions which predict kernel weights, DCNs learn to dynamically transform the sampling location of the otherwise fixed convolution filters.

\subsubsection{Pre-training} of backbone networks is commonly done as supervised learning on ImageNet (IN). Due to the recent success of self supervised learning (SSL) in classification, we are interested in how these representations perform in terms of object-centric robustness. In particular we are interested in the contrastive learning framework that seeks to learn ``representations with enough invariance to be robust to inconsequential variations \cite{tian_what_2020}''. Based on availability we include the methods InstDis \cite{wu_unsupervised_2018}, MoCo \cite{he_momentum_2020,chen_improved_2020}, PIRL \mbox{\cite{misra_self-supervised_2020}} and InfoMin \cite{tian_what_2020}. Note that pre-trained backbones were only used as initialization for a supervised training on COCO. As a final comparison we include models that have been trained with random initialization and Large Scale Jittering (LSJ) \cite{ghiasi_simple_2021} data augmentation as an alternative to pre-training.
\newline

\begin{table}[t]
   \caption{Overview of frameworks, backbones and neck methods. (*) Swin Transformer use hierarchical representations similar to FPN necks in CNNs. RegNetY is similar to RegNetX but implements the Squeeze-and-Excitation operation \cite{hu_squeeze-and-excitation_2018}. Yolo consists of darknet (D), spatial pyramid pooling (SSP) \cite{fleet_spatial_2014} and a Path Aggregation Network (PAN) \cite{liu_path_2018-1} in varying combinations with CSPNet (C) \cite{wang_cspnet_2020}.}
   \label{tab:model-overview}
   \begin{center}
   \resizebox{0.9\textwidth}{!}{
   \begin{tabular}{ |c||l|l|l|l|l|l|l|l|l| }
      \multicolumn{1}{c}{
         \textbf{Backbone}}   & \multicolumn{9}{c}{\textbf{Framework}}                                                                                                                                                                \\ \hline
                           CNN    & \multicolumn{3}{c|}{multi stage}                          & \multicolumn{6}{c|}{one stage}                                                                                                        \\ \cline{2-10}
      \cellcolor{gray!20}  GCN    & \multicolumn{5}{c|}{anchor based}                                                      & \multicolumn{4}{c|}{anchor free}                                                                         \\ \cline{2-10}
      \cellcolor{green!20} Hybrid & \multicolumn{7}{c|}{top down (bbox\(\to\)segm)}                                                                                               & \multicolumn{2}{c|}{bottom up (segm\(\to\)bbox)}  \\ \cline{2-10}
      \cellcolor{blue!20}  ViT    & Mask R-CNN              & BMask & Cascade                 & YOLO(v3,4,s4) & YOLACT(++) & DETR                         & FCOS BCNet             & SOTR                          & SOLOv2           \\ \hline \hline
      R50                         & FPN, C4, DC5, DCN       & FPN   & FPN                     & -             & FPN, DCN   & \cellcolor{green!20}FPN, DC5 & -                      & -                             & FPN              \\
      R101                        & FPN, C4, DC5,           & FPN   & DCN                     & -             & FPN, DCN   & \cellcolor{green!20}FPN      & \cellcolor{gray!20}FPN & \cellcolor{green!20} FPN, DCN & FPN              \\
      X101                        & FPN                     & -     & -                       & -             & -          & -                            & -                      & -                             & -                \\
      X151                        & -                       & -     & FPN, DCN                & -             & -          & -                            & -                      & -                             & -                \\
      RegNetX                     & FPN                     & -     & -                       & -             & -          & -                            & -                      & -                             & -                \\
      RegNetY                     & FPN                     & -     & -                       & -             & -          & -                            & -                      & -                             & -                \\
      Swin-T                      & \cellcolor{blue!20}FPN* & -     & \cellcolor{blue!20}FPN* & -             & -          & -                            & -                      & -                             & -                \\
      Swin-S                      & \cellcolor{blue!20}FPN* & -     & \cellcolor{blue!20}FPN* & -             & -          & -                            & -                      & -                             & -                \\
      Swin-B                      & -                       & -     & \cellcolor{blue!20}FPN* & -             & -          & -                            & -                      & -                             & -                \\ 
      D53                         & -                       & -     & -                       & FPN           & -          & -                            & -                      & -                             & -                \\ 
      CD53                        & -                       & -     & -                       & (C)PAN, SPP   & -          & -                            & -                      & -                             & -                \\ \hline
   \end{tabular}}
   \end{center}
\end{table}

From the overview in \autoref{tab:model-overview} we can now derive dimensions that allow for a fair evaluation of models. Specifically, we can fix the backbone and neck architecture (e.g. ResNet + FPN) for a controlled comparison between \textit{frameworks}. Vice versa, we can investigate the impact of a specific backbone and neck combination within a fixed framework (e.g. Mask R-CNN). The complete list of models is displayed in \autoref{fig:model-overview-ap}. Note that we did not include the dimension of pre-training in the above overview for readability reasons. In our experiments however, we do compare training setups and learning schedules for a fixed model type \mbox{(e.g. Mask R-CNN + ResNet + FPN)}.

\section{Related Work}

Robust vision can be approached from different perspectives. The classical view stems from signal processing and concerns image corruptions that are independent from the signal, e.g. salt and pepper noise \cite{hendrycks_benchmarking_2019,michaelis2019dragon,kamann_benchmarking_2021,mummadi_does_2021}. A popular alternative is to compare model performance and failure cases against humans \cite{geirhos_generalisation_2018,geirhos_imagenet-trained_2019,geirhos_surprising_2020,geirhos_partial_2021,shankar_evaluating_2020,tuli_are_2021}. Since human vision is fairly robust, the hope is that vision models with a similar bias will generalize in more systematic ways as well. For instance, \cite{geirhos_partial_2021} and \cite{tuli_are_2021} show that transformer models perform closer to human behavior than CNNs. In support of the hypothesis, \cite{mahmood_robustness_2021} find that transformer architectures are more robust against adversarial attacks. \cite{madan_small_2021} on the other hand report that both model types are prone to small in-distribution changes in 3D perspective and lighting. The approach we take is inspired by these works but more direct. Instead of comparing to humans, it measures texture robustness in a challenging zero-shot setting. A third perspective originates from the long-tail distribution of real-world data. In such settings, robustness can be understood as the ability to adapt to uncommon or novel objects with efficient transfer learning \cite{hu_learning_2018-1,vedaldi_commonality-parsing_2020}, re-sampling \cite{vedaldi_devil_2020,chang_image-level_2021} or regularization strategies \cite{pan_model_2021,hsieh_droploss_2021,wang_seesaw_2021}. Particular relevant are methods that handle object occlusion \cite{chen_multi-instance_2015}. Since objects can be occluded in almost infinite ways, a common strategy is to represent object properties more explicitly, e.g. to decoupled shape and appearance for instance \cite{vedaldi_boundary-preserving_2020,ke_deep_2021,vedaldi_commonality-parsing_2020}. We expect these methods to be strong contenders in our comparison.
\newline

As an alternative to our sensitivity benchmark, \cite{islam_shape_2021} analyze feature importance in latent representations and \cite{cao_inverting_2021} use feature visualization to understand object detectors. Both leverage style transfer to simulate novel object appearances. The closest real-life alternative is the Natural Adversarial Objects (NAO) dataset \cite{lau_natural_2021}. It depicts a more realistic out-of-distribution setting but does not allow to control for pose and perspective, i.e. to observe the exact same objects with varying textures for instance. Other alternatives are the 3DB framework \cite{leclerc_3db_2021}, a rendering engine that enables artifact free texture transfer on synthetic objects and SI-Score \cite{yung_si-score_2021}, a dataset for analyzing robustness to rotation, location and size.

\section{Results}

\begin{figure}[h!]
\begin{center}
\includegraphics[trim=0 35 0 0, width=\textwidth]{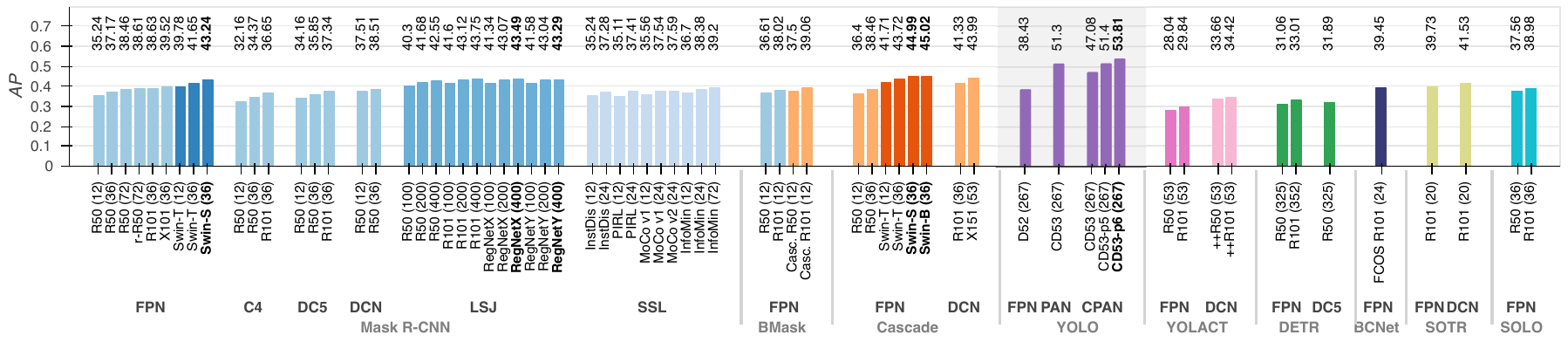}
\end{center}
\caption{Absolute performances on COCO \texttt{val2017}. Training schedules in epochs have been appended to model names. Note that Yolo is bounding box AP which is not comparable but included for model completeness. Methods that did not report scores for \texttt{val2017} have been validated on \texttt{test-dev2017} first}
\label{fig:model-overview-ap}
\end{figure}

In this section we present the zero-shot evaluation on Stylized COCO, Objects and Background. Each dataset version contains 20 copies of the accordingly styled COCO \texttt{val2017} subset. As a reference point, we \textit{reproduce the evaluation} on the original \texttt{val2017} subset and report the absolute Average Precision (AP) for all models in \autoref{fig:model-overview-ap}. As can be seen, training schedule, data augmentation and architecture choice have the biggest impact within a framework. Overall, RegNets trained with LSJ and Swin Transformer models perform best. Note that SOTR and SOLO have worse APs but significantly better APm and APl compared to other frameworks (see supplementary material for all scores).
\newline

We now present the results of our sensitivity analysis. In total, we tested 68 pre-trained models \footnote{\tiny See supplementary material for the list of code projects and weight sources.} on 61 \texttt{val2017} replicas which sums up to 4148 subset evaluations. Since it is not expedient to report this amount of data in the form of tables, we communicate mainly with figures in the main paper. However, the exact numerical values will be released for inspection together with the code. To quantify out-of-distribution robustness, we calculate the relative zero-shot performance in comparison to the performance on uncorrupted data. For every dataset version, blending space and alpha step we calculate:
\begin{equation}
rP_{\alpha} = P_{\alpha} / P_{coco}; P \in \{AP, APs, APm, APl\}
\end{equation}
Note that we focus on IoU type segmentation and the scale dependent \(APs,m,l\) metrics since COCO has an unbalanced distribution of \(41\%\) small, \(34\%\) medium and \(24\%\) large instances. We include more metrics, absolute scores and the corresponding figures for bounding box IoU in the supplementary material.

\subsubsection{A large scale comparison} 
\begin{figure}[t]
   \begin{center}
   \includegraphics[width=0.74\textwidth]{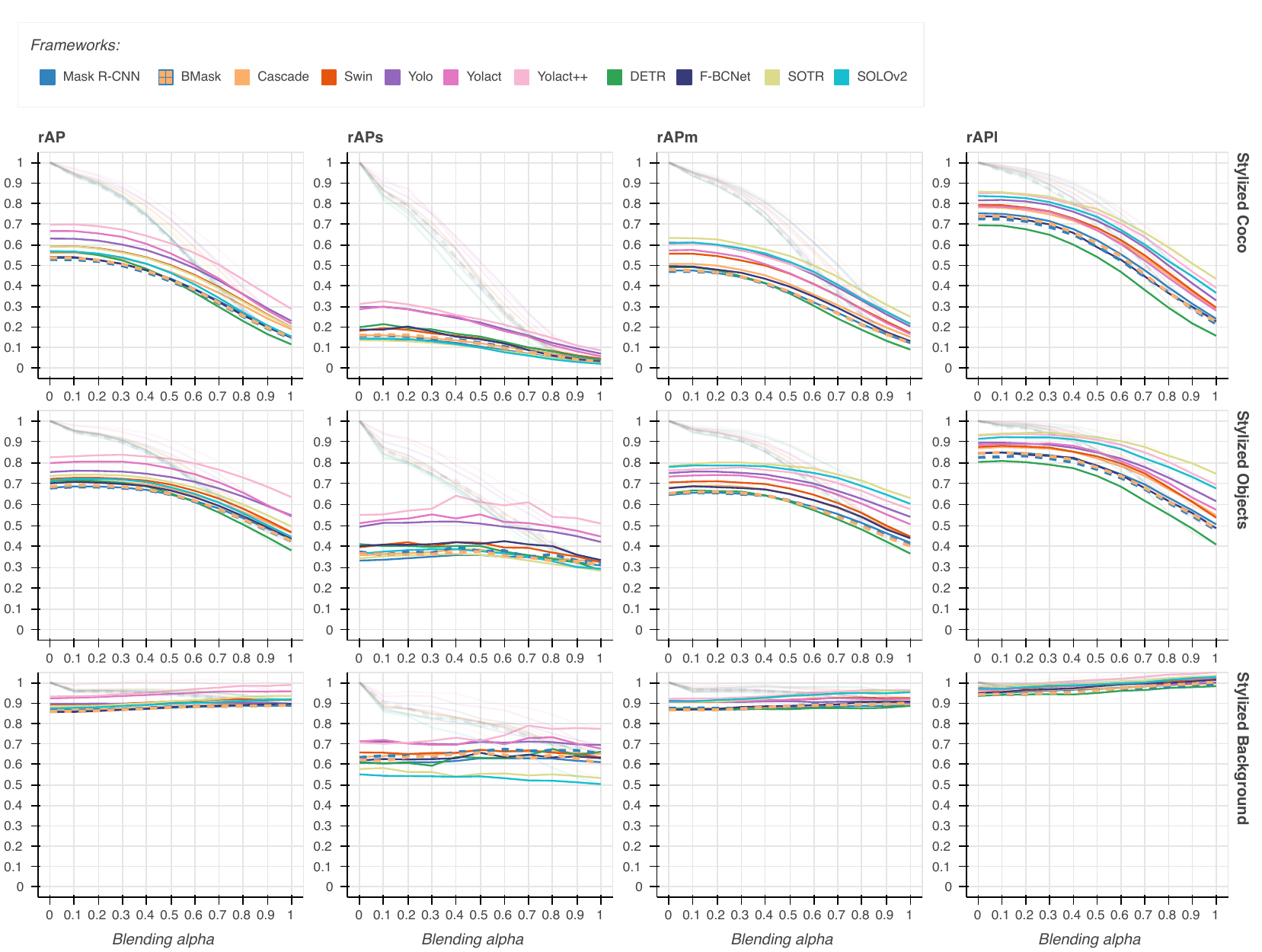}
   \caption{General overview of model robustness against out-of-distribution texture (zoom in for better visibility). Yolo is displayed for completeness but not directly comparable.}
   \label{fig:large-comparison}
   \end{center}
\end{figure}

of all models can be found in \autoref{fig:large-comparison}. Relative AP scores are displayed from left to right, datasets from top to bottom. The subfigures can be read as follows. From left to right (\(\alpha:  0 \to 1\)), how much of the original performance is lost with increasing out-of-distribution texture? Recall that at \(\alpha = 0\), no out-of-distribution texture is used. For the feature space sequence (dark colors), this means that a loss in performance can be, at this point, attributed to image corruptions from the style transfer method. For the pixel space control sequence (light colors), this point depicts the original \texttt{val2017} score which is always \(100\%\) in our relative metric. For better visibility, we display an averaged model group per framework. This decision is not arbitrary however. Between all models, we calculated the average L2 distance over the full alpha range (see supplementary material for the resulting distance matrices). As a result of this analysis, we find models from the same framework to perform more similar to each other (\(\mu_{L2} = 0.08\pm0.05\)) than to models from other frameworks (\(\mu_{L2} = 0.21\pm0.08\)). Note that Swin Transformer is treated as a custom \textit{framework} in this comparison even though it implements the (Cascade) Mask R-CNN strategies. In the following, we highlight a few key observations.
\newline

\textit{First subfigure (top left):} The average framework AP on Stylized COCO starts from only \(58\pm5\%\) due to the impact of image corruption. Revisit \autoref{fig:dataset-details} to see how images look like at this point (\(\alpha=0\)). After this initial loss, performance remains fairly constant up to \(\alpha=0.3\) from where it drops to \(18\pm5\%\). Apparently, frameworks seem to maintain a consistent ranking over the full alpha range which we investigate in more detail in our controlled comparison.
\newline

\textit{First row (Stylized COCO):} On small objects (rAPs), model performance is severely affected by image corruptions from the beginning: \(19\pm6\% \to 4\pm2\%\). With increasing objects size, this effect seems to gradually vanish. On large objects (rAPl), the relative performance is then more affected by actual out-of-distribution texture than by corruption artifacts: \(78\pm5\% \to 28\pm8\%\).
\newline

\textit{Second row (Stylized Objects):} The average framework model is more robust to this type of data, e.g. rAP is now: \(72\pm5\% \to 47\pm7\%\). In addition, models exhibit an extended range of robustness after the initial dip (now up to \(\alpha=0.5\)). The apparent ranking of frameworks remains the same. Note the increasing variance towards the end of the blending sequence, e.g. rAPm: \(70\pm5\% \to 48\pm9\%\), rAPl: \(87\pm4\% \to 56\pm10\%\). We conclude that object contour is in fact an important property and that some models can exploit this feature more effectively than others. However, no model is able to exploit the implicit encoding of ground truth masks. Doing so would imply a high level of abstraction which we do not find to happen in existing models.
\newline

\textit{Third row (Stylized Background):} In this dataset version, only background features are corrupted. This allows us to measure the importance of context information. As displayed in the second subfigure (rAPs), models do indeed rely on additional context information to segment small objects: \(63\pm5\% \to 63\pm7\%\). With increasing object size, models depend less on background information. Interestingly, we observe a subtle performance gain on large objects with increasing stylization (rAPl): \(96\pm2\% \to 101\pm2\%\). As before however, models do not exploit the implicit mask encoding in a more systematic way which is arguably even easier in this setting.

\subsubsection{A controlled comparison} of frameworks, architectures and pre-training is presented in this subsection. Each comparison contributes to the main goal of our study which asks: \textit{Do existing segmentation methods contain components or design decisions that promote systematic generalization regarding object-centric features?} Note that we focus on the extreme points \(\alpha=0\) and \(\alpha=1\).\newline

\begin{figure}[t]
   \begin{center}
   \includegraphics[trim=0 35 0 0,width=\textwidth]{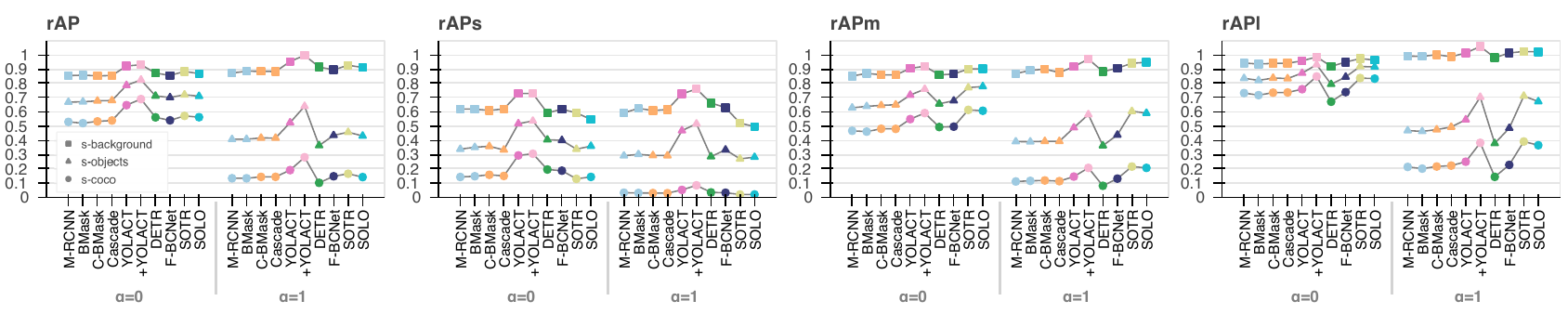}
   \end{center}
   \caption{A controlled comparison of framework robustness. Note that we compromise on ResNet-101 for SOTR and F-BCNet, see \autoref{tab:model-overview} for available backbones}
   \label{fig:frameworks}
\end{figure}

\textit{Frameworks} are compared first. For statistical fairness, we use the same backbone and neck method per framework (ResNet-50 + FPN). \autoref{fig:frameworks} displays the results on the feature space sequence. As hypothesized earlier, we observe a consistent ranking between frameworks from \(\alpha: 0 \to 1\) and across dataset versions. A noticeable difference is the amplified difference between frameworks and the increased distance between Stylized COCO and Stylized Objects at \(\alpha = 1\). We conclude that robustness against semantic image corruption is correlated with robustness against out-of-distribution texture. However, some frameworks seem to exploit object contour more effectively than others. The most robust framework overall is YOLACT++. Its predecessor YOLACT is compatible on small objects. SOTR and SOLO(v2) perform best on medium and large objects. DETR appears as the least robust framework but we like to point out that it was not primarily designed for instance segmentation. Surprisingly, BMask and BCNet are not better than Cascade and Mask R-CNN.\newline

\begin{figure}[h]
   \begin{center}
   \includegraphics[trim=0 35 0 0,width=\textwidth]{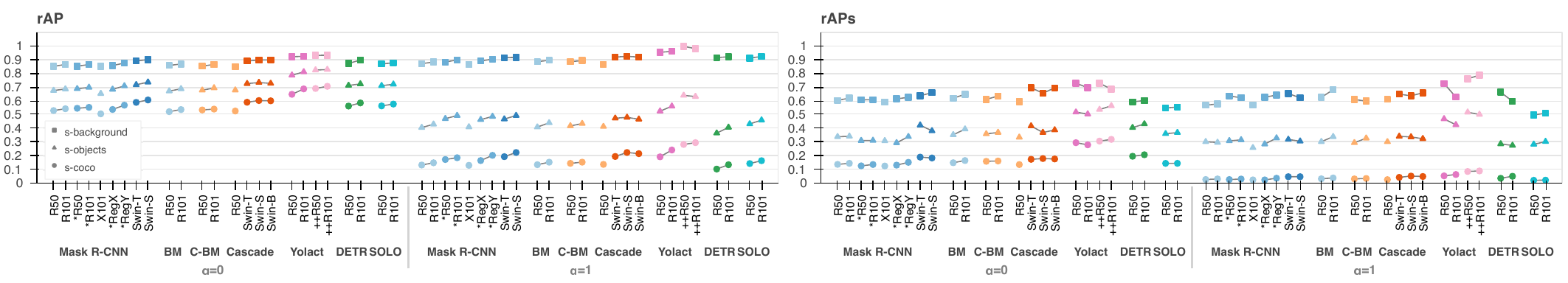}
   \end{center}
   \caption{Model robustness for different backbones. Scores for medium and large objects follow the trend of rAP and can be found in the supplementary material. Models marked with * are trained with LSJ}
   \label{fig:backbones}
\end{figure}

\textit{Backbones} are compared next. In this analysis, we only use models with FPN neck and compare within the same framework. When available, we include model pairs with different data augmentation. The results are displayed in \autoref{fig:backbones}. As can be seen from the left subfigure (rAP), a clear trend is visible. When every other factor is controlled, deeper backbones improve robustness. Swin-T and S are comparable to ResNet-50 and 101 in model complexity but generally more robust than their CNN counterparts. For medium and large objects, a similar trend can be reported. For small objects, the behavior is less clear and surprisingly, often reversed.

\begin{figure}[h]
   \begin{center}
   \includegraphics[trim=0 35 0 0, width=\textwidth]{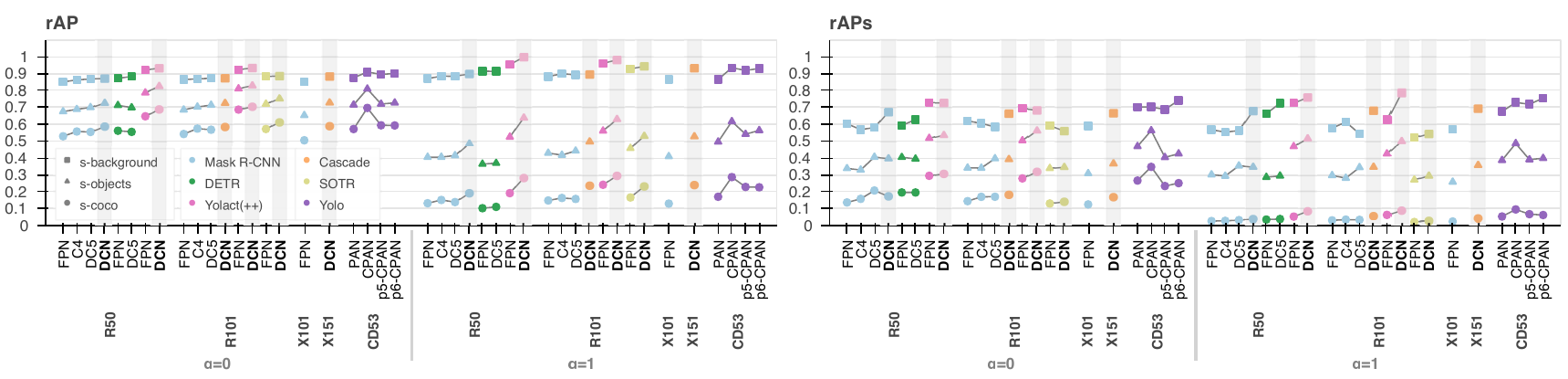}
   \end{center}
   \caption{Robustness by neck architecture. Deformable convolutions are highlighted for better visibility. Note that Yolo is only comparable to itself due to bounding box score}
   \label{fig:neck}
\end{figure}

\textit{Neck methods} are compared in \autoref{fig:neck}. We only use models with a similar training schedule and group them by backbone and framework. Surprisingly, the popular FPN method is consistently the least robust option. In contrast, deformable convolutions (DCN) consistently improve robustness as highlighted for better visibility. For small objects, DC5 necks improve the robustness on Stylized Objects (triangle). Note that Yolo is only comparable against itself with the simple \mbox{v4-csp} model (CPAN) performing best.

\begin{figure}[h]
   \centering
   \includegraphics[width=0.3818\textwidth]{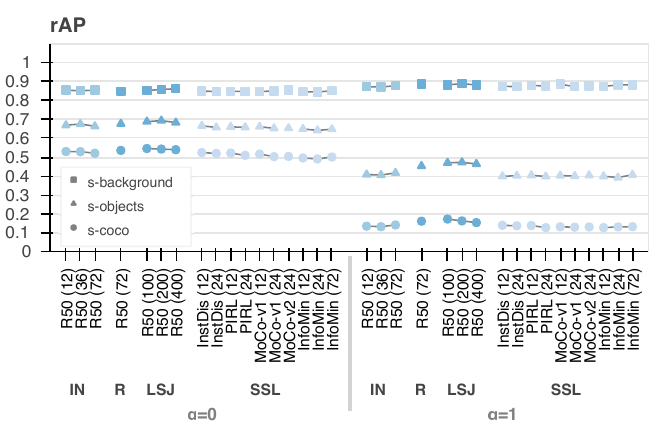}
   \caption{Model robustness by pre-training and data augmentation. We distinguish between ImageNet (IN), random (R) and self supervised learning (SSL) initialization}
   \label{fig:pre-training}
\end{figure}

\textit{Pre-training, training schedules and data augmentation} have surprisingly, little to no impact on texture robustness. The result of this comparison is displayed \autoref{fig:pre-training}. On the other hand we can report that supervised and unsupervised representations perform almost on par. We suspect that the supervised fine tuning on the segmentation task overshadows potential differences in the initial backbone representations. In comparison we find that random initialization with LSJ data augmentation is the most robust combination.

\section{Discussion}

As expected, we found deep learning models for instance segmentation to be vulnerable to image corruption and not particularly robust against novel object texture. Based on our in depth analysis, we now understand this problem space much better. In particular, we understand that the type of robustness depends heavily on object size. For instance, small objects are predominantly affected by image corruption. In consequence, we hypothesize that the difference between shape and texture simply collapses for (very) small objects in MS COCO. Furthermore, we find that object contour is an important feature to segmentation models that can be exploited independently from object texture (better performance on Stylized Objects). However, the question arises if this is not simply a byproduct of the limited area of effect in this dataset. To answer this question, we conduct an ablation study with salt and pepper noise which rejects this hypothesis. The results can be found in the supplementary material. For medium and large objects, we found models to be biased towards learned texture, i.e. out-of-distribution texture results in a significant performance loss. In contrast to \cite{geirhos_imagenet-trained_2019} who argue that this problem is induced by the ImageNet training data, we find no such correlation in terms of pre-training for instance segmentation. We conclude that either a similar bias is induced by the COCO dataset or that architecture choice is simply more important as suggested by \cite{hendrycks_natural_2021}. Our key findings support the latter hypothesis. In particular, the promising results of YOLACT(++), SOTR and SOLOv2 suggest more in depth research. In the same spirit, \cite{greff_binding_2020} formulate \textit{the binding problem} which is ``the inability [of neural networks] to dynamically and flexibly combine (bind) information [... which] limits their ability to [...] accommodate different patterns of generalization''. Our results for dynamic architectures confirm this assumption to the extend that we consider this another promising research direction.
\newline

\vspace{-0.5cm}
\section{Conclusion}

In this study we contribute a comprehensive baseline on the texture robustness of deep learning based instance segmentation. As a result of our study, we find a noticeable texture bias in most existing methods. However, models do also exploit other features such as object contour. Based on this insight we feel optimistic that, with the right design decisions, vision models are not trapped in texture bias. The key finding of our study is that the frameworks YOLACT++, SOTR and SOLOv2 as well as deeper and dynamic architectures improve model robustness. We hope that our rigorous analysis enables more in depth research along these lines and more generally, a systematic design of robust vision models.

\clearpage
%
%
\bibliographystyle{splncs04}
\bibliography{ms}
\end{document}